\documentclass[letterpaper, 10 pt, journal, twoside]{ieeetran} 

\usepackage{microtype}
\usepackage{graphicx}
\usepackage{booktabs} 

\usepackage{hyperref}

\usepackage{bbold}
\usepackage{tcolorbox}
\newtcolorbox{mybox}[3][]
{
  colframe = #2!25,
  colback  = #2!10,
  coltitle = #2!20!black,  
  title    = {#3},
  #1,
}
\definecolor{Q_rgb}{RGB}{66,115,177}
\definecolor{A_rgb}{RGB}{157,31,58}
\newcommand{\VarSty}[1]{\textnormal{\ttfamily\color{purple!90!black}#1}\unskip}

\definecolor{linkc}{rgb}{0, 0.44, 0.74}
\definecolor{eqc}{rgb}{1, 0, 0}

\usepackage{xr}
\definecolor{mygreen}{RGB}{69, 191, 85}
\definecolor{myred}{RGB}{238, 75, 43}
\usepackage{booktabs}
\usepackage{multirow}
\usepackage{graphicx}
\usepackage{url}
\usepackage{wrapfig,lipsum,booktabs}
\usepackage{amsmath,amssymb,stmaryrd} 
\usepackage[font=small]{caption, subcaption}

\newcommand\mypara[1]{\vspace{0mm}\noindent\textbf{#1}}

\title{DriveGPT4: Interpretable End-to-end \\Autonomous Driving via Large Language Model}
\author{Zhenhua Xu, 
Yujia Zhang, Enze Xie*, Zhen Zhao, Yong Guo, \\Kwan-Yee K. Wong, Zhenguo Li, Hengshuang Zhao*  %

\thanks{Manuscript received April 2, 2024; Revised June 11, 2024; Accepted July 9, 2024. This paper was recommended for publication by Editor Abhinav Valada upon evaluation of the Associate Editor and Reviewers' comments. This work is supported by the National Natural Science Foundation of China (No. 62201484), HKU Startup Fund, and HKU Seed Fund for Basic Research.}
\thanks{Zhenhua Xu, Kwan-Yee K. Wong, Hengshuang Zhao are with The University of Hong Kong (email: zxubg@connect.ust.hk).}
\thanks{Yujia Zhang is with the Zhejiang University.}
\thanks{Enze Xie, Yong Guo, Zhenguo Li are with Huawei Noah's Ark Lab.}
\thanks{Zhen Zhao is with University of Sydney.} 
\thanks{\textit{(Corresponding author: Enze Xie, Hengshuang Zhao})} 
\thanks{Digital Object Identifier (DOI): see top of this page.} 
} 

\IEEEaftertitletext{\vspace{-1.0\baselineskip}}
\begin{document}
\bstctlcite{IEEEexample:BSTcontrol}
\maketitle

\begin{abstract}

Multimodal large language models (MLLMs) have emerged as a prominent area of interest within the research community, given their proficiency in handling and reasoning with non-textual data, including images and videos. This study seeks to extend the application of MLLMs to the realm of autonomous driving by introducing DriveGPT4, a novel interpretable end-to-end autonomous driving system based on LLMs. Capable of processing multi-frame video inputs and textual queries, DriveGPT4 facilitates the interpretation of vehicle actions, offers pertinent reasoning, and effectively addresses a diverse range of questions posed by users. Furthermore, DriveGPT4 predicts low-level vehicle control signals in an end-to-end fashion.
These advanced capabilities are achieved through the utilization of a bespoke visual instruction tuning dataset, specifically tailored for autonomous driving applications, in conjunction with a mix-finetuning training strategy.  DriveGPT4 represents the pioneering effort to leverage LLMs for the development of an interpretable end-to-end autonomous driving solution. Evaluations conducted on the BDD-X dataset showcase the superior qualitative and quantitative performance of DriveGPT4. Additionally, the fine-tuning of domain-specific data enables DriveGPT4 to yield close or even improved results in terms of autonomous driving grounding when contrasted with GPT4-V. The webpage of this paper is available at \url{https://tonyxuqaq.github.io/projects/DriveGPT4}.
\end{abstract}

\section{Introduction}
Over the past decade, there has been remarkable growth in the field of autonomous driving, encompassing both academia and industry \cite{liu2021role,parekh2022review}. Commercialized autonomous driving systems have been successfully implemented in everyday scenarios, such as harbors, warehouses and urban areas. Commonly, the autonomous vehicle adopts modular designs, including perception, planning, and control. In conventional autonomous driving systems, these modules are implemented by detailed rule-based methods to handle various scenarios. But such a system may fail when unseen cases are met, such as rare accidents.

To ensure that vehicles can effectively handle diverse situations using intelligent actions, data-driven learning-based methods have become a widespread component of modern autonomous driving systems \cite{zhao2017pyramid,xue2019learning,xu2023centerlinedet,xu2023rngdet++,xu2023insightmapper}. To better integrate and optimize the entire system, some approaches propose training the network in an end-to-end manner, eliminating the need for discontinuous intermediate steps \cite{prakash2021multi,hu2023planning,chen2023end}. By using vehicle-mounted sensor data as input, the end-to-end autonomous driving system can directly predict planned paths and/or low-level vehicle controls. 
Nonetheless, the end-to-end learning-based autonomous driving system functions as a black box, signifying that humans cannot interpret or comprehend the generated decisions, leading to significant ethical and legal concerns, which restricts the development of commercialized autonomous driving systems.

In recent years, explainable autonomous driving \cite{deruyttere2019talk2car,kim2019CVPR,atakishiyev2021explainable,jin2023adapt,malla2023drama} has garnered increasing interest due to its potential to demystify the black box. These studies develop large-scale datasets comprising autonomous vehicle data along with language pairs.
Language models, such as BERT \cite{devlin2018bert} and GPT \cite{radford2018improving}, are trained on these datasets to generate natural language based on input from vehicle-mounted sensor data.  
However, the capabilities of small language models are limited, causing most of these systems to produce rigid responses to predefined questions. In addition, small language models suffer from insufficient model capacity and present unsatisfactory question-answering performance.

With the advent of large language models (LLMs), such as ChatGPT \cite{chatgpt2023} and LLaMA \cite{touvron2023llama}, the interpretability of autonomous driving systems could benefit from improved text prediction, given that LLMs possess extensive general knowledge about the world. Moreover, LLMs have the potential to better analyze and generate low-level vehicle controls due to their inherent reasoning capabilities. To achieve this, LLMs are required to comprehend multimodal data, like images or videos. Multimodal LLMs have been attracting increasing interest from various research communities, such as computer vision \cite{li2022grounded,li2022blip}, embodied AI \cite{driess2023palm,liang2023code}, and biomedicine \cite{karabacak2023embracing,li2023llava}. These studies propose to project multimodal input from image, audio, video, control, and other spaces into the text domain, allowing LLMs to understand and process this multimodal data as text. To the best of our knowledge, no existing paper grounds LLMs for interpretable end-to-end autonomous driving purposes. 

In this paper, we introduce DriveGPT4, an interpretable end-to-end autonomous driving system that utilizes large language models. The digit ``4'' in the system name represents multimodality, similar to that of MiniGPT4 \cite{zhu2023minigpt}. DriveGPT4 takes as input a video sequence captured by a front-view monocular RGB camera, and then predicts the control signal for the next step (i.e., vehicle speed and turning angle). At the same time, human users can converse with DriveGPT4, which can provide natural language responses, such as describing the vehicle's actions and explaining the reasoning behind its behavior. To train DriveGPT4 to communicate like a human, we follow LLaVA \cite{liu2023visual} and create a visual instruction tuning dataset based on the BDD-X dataset \cite{kim2018textual} using ChatGPT. The contributions of this paper are summarized as follows:

\begin{itemize}
    
    \item We present DriveGPT4, a novel multimodal LLM for interpretable end-to-end autonomous driving. Mix-finetuned on the created dataset, DriveGPT4 can process multimodal input data and generate text responses as well as low-level control signals.
    \item We develop a new visual instruction tuning dataset for interpretable autonomous driving with the assistance of ChatGPT. The performance of DriveGPT4 is boosted by finetuning the generated data. 
    \item We evaluate all methods on the BDD-X dataset for multiple tasks. DriveGPT4 outperforms all baselines, which demonstrates its effectiveness. 
\end{itemize}

\section{Related Works}
\mypara{End-to-end Autonomous Driving.}
End-to-end autonomous driving aims to directly predict the vehicle path and low-level control signals based on visual inputs \cite{bojarski2016end,xiao2020multimodal,prakash2021multi,hu2023planning,chen2023end}. \cite{he2016deep} is considered the first deep learning end-to-end self-driving work. In this study, the authors train a convolutional neural network to control vehicles using monocular images as input. Recent works integrate all system modules by tokenizing module outputs \cite{hu2023planning,chen2023end}, achieving a more powerful and robust control effect. However, these works lack interpretability, which limits their trustworthiness and commercialization potential.

\mypara{Interpretable Autonomous Driving.}
To address the black box issue in learning-based autonomous driving, some studies employ visualizations \cite{kim2017interpretable,wang2021learning,saha2022translating}. However, visual maps can be challenging for non-expert passengers to comprehend. Alternatively, other research utilizes language models to describe vehicle situations with natural languages, such as vehicle actions \cite{deruyttere2019talk2car,kim2019CVPR,jin2023adapt}, vehicle action reasoning \cite{jin2023adapt}, surrounding object statements \cite{malla2023drama}, and discussions of potential risks to the ego vehicle \cite{malla2023drama}. Constrained by the limited capacity of smaller language models, these methods can only address predefined human questions and provide inflexible answers, hindering their widespread application in real-world scenarios.

\mypara{Multimodal LLM.}
 Building on the powerful pretrained LLM weights, such as PaLM \cite{chowdhery2022palm,driess2023palm}, LLaMA \cite{touvron2023llama,touvron2023llama2}, and Vicuna \cite{peng2023instruction}, multimodal LLMs aim to handle multiple types of input beyond text. Blip \cite{li2022blip,li2023blip} leverages Q-formers to project multimodal input into the text space, while others \cite{li2023llava,luo2023valley} simply train a fully connected layer as the projector. Multimodal LLMs have been widely applied to various tasks, such as image understanding \cite{li2023blip,liu2023visual}, video understanding \cite{luo2023valley,zhang2023video,wang2023chatvideo,zhu2023minigpt,li2023videochat}, medical diagnosis \cite{li2023llava,karabacak2023embracing}, and embodied AI \cite{chowdhery2022palm,driess2023palm,brohan2023rt,liang2023code}, etc. Our task is closely related to video understanding and embodied AI. DriveGPT4 is inspired by the former to understand input video data and the latter to predict control signals. Among these works, only a few focus on autonomous driving-related tasks \cite{fu2023drive,wu2023language,sima2023drivelm}. DriveLikeHuman \cite{fu2023drive} can only handle simple simulation scenes, limiting its real-world applicability. NuPrompt \cite{wu2023language} focuses on object tracking for vehicle perception but does not consider end-to-end driving or vehicle action reasoning. DriveLM \cite{sima2023drivelm} is a large benchmark for driving scene understanding. 
 

\begin{figure}[t]
  \centering
    \includegraphics[width=\linewidth]{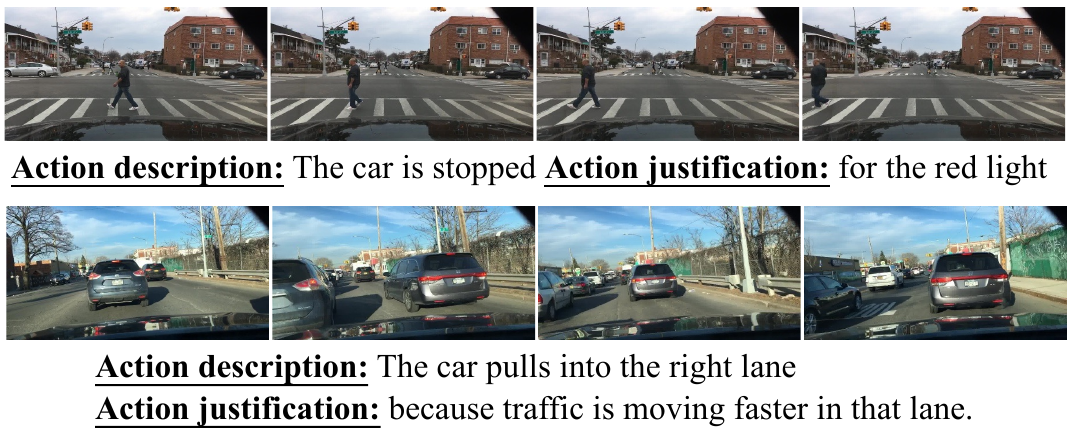}
  \caption{Example of BDD-X labeled data.}
  \label{bddx}
\vspace{-10pt}
\end{figure}
\begin{figure*}[t]
  \centering
    \includegraphics[width=\linewidth]{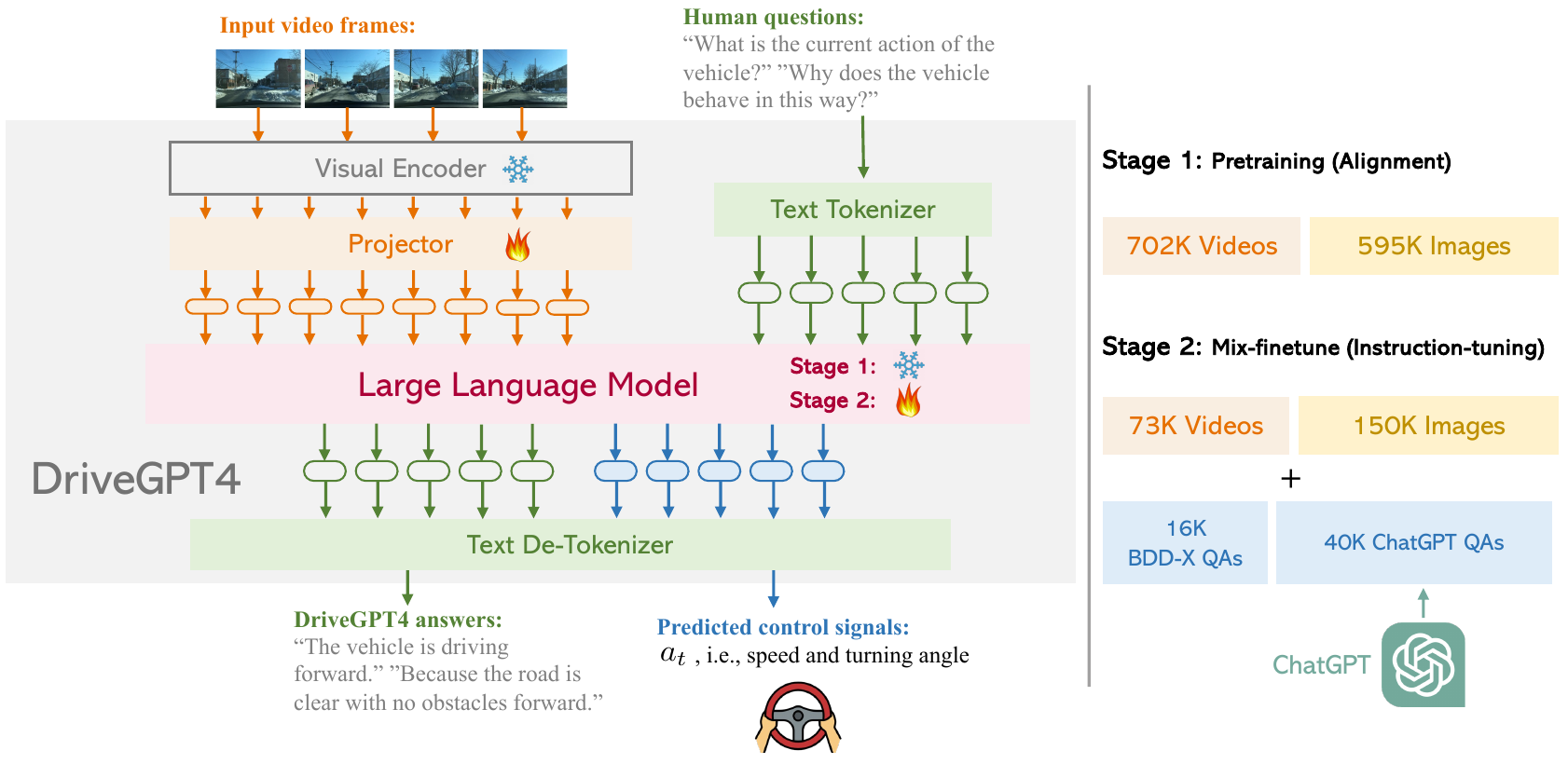}
  \caption{DriveGPT4 overview. DriveGPT4 is a comprehensive multimodal language model capable of processing inputs comprising videos, and texts. Video sequences undergo tokenization using a dedicated video tokenizer, while text and control signals share a common de-tokenizer. DriveGPT4 can concurrently generate responses to human inquiries and predict control signals.}
  \label{diagram}
\vspace{-15pt}
\end{figure*}

\section{Data Generation}\label{data_generation}
\subsection{BDD-X Dataset.} The BDD-X dataset \cite{kim2018textual} is employed in this study due to the scarcity of publicly available datasets suitable for our task. We sourced both videos and labels from the BDD-X dataset. This dataset contains approximately 20,000 samples, which consist of 16,803 clips designated for training and 2,123 for testing. Each clip is divided into eight images. The BDD-X dataset provides control signal data for each frame, such as vehicle speed and turning angle. It also includes text annotations detailing vehicle action descriptions and action justifications for every video clip, as exemplified in Fig. \ref{bddx}.

\mypara{BDD-X question-answerings.} BDD-X provides three types of labels: vehicle action descriptions, action justifications, and control signals for each video clip. To train the LLM, question-answering (QA) pairs are required. We generate a set of synonymous questions and use corresponding BDD-X labels as the answer. For example, for a vehicle action description,  a question equivalent to \textit{``What is the current action of this vehicle?"} should be sent to the LLM as the input question. Then, the LLM should generate the response, whose ground truth label is the vehicle action description. Considering there are three types of labels in the BDD-X dataset, we create three question sets: $Q_a$, $Q_j$, and $Q_c$. To prevent the LLM from overfitting to fixed question patterns, inspired by \cite{liu2023visual}, each question set should contain multiple synonymous expressions of one question.

\begin{itemize}
    \item $Q_a$ contains synonymous questions equivalent to \textit{``What is the current action of this vehicle?"}. 
    A randomly selected question $q_a\in Q_a$ forms a QA pair with the action description label.
    \item $Q_j$ contains synonymous questions equivalent to \textit{``Why does this vehicle behave in this way?"}. 
    A randomly selected question $q_j\in Q_j$ forms a QA pair with the action justification label.
    \item $Q_c$ contains synonymous questions equivalent to \textit{``Predict the speed and turning angle of the vehicle in the next frame."}. 
    A randomly selected question $q_c\in Q_c$ forms a QA pair with the control signal label.
\end{itemize}
A randomly selected question $q_X\in Q_X$ and a corresponding label form a QA pair to create the dataset.
LLMs can learn to predict and interpret vehicle actions simultaneously. However these QA pairs have fixed and rigid contents. Due to the lack of diversity, training solely on these QAs will degrade the ability of LLMs and render them incapable of answering questions in other formats.

\begin{table*}[t!]\centering
\begin{minipage}{2.0\columnwidth}\vspace{0mm}    \centering
\begin{mybox}{purple}
    \centering
   
      \footnotesize
    \begin{tabular}{p{0.97\columnwidth} c}
    
    \includegraphics[height=2cm]{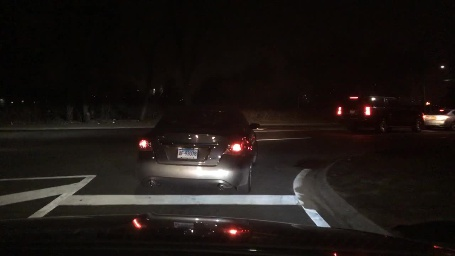}\hspace{3pt}\includegraphics[height=2cm]{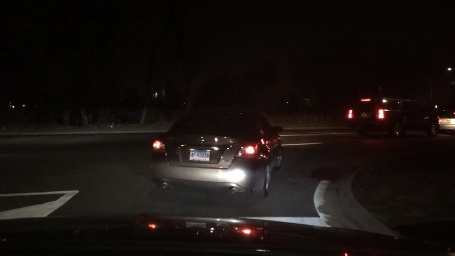}\hspace{3pt}\includegraphics[height=2cm]{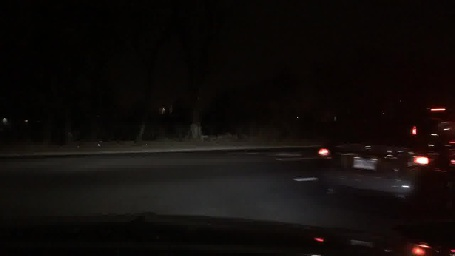}\hspace{3pt}\includegraphics[height=2cm]{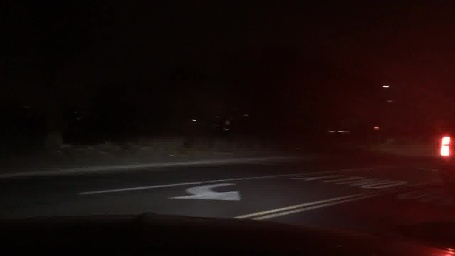}\\
   \VarSty{ {\bf Caption1: Action description} } &\\
The car turns right&\\

\VarSty{ {\bf Caption 2: Action justification} } & \\
As the road is clear to turn.&\\
\VarSty{ {\bf Control signals} } & \\
Speed(m/s): 3.91, 3.1, 2.35, 2.92, 3.51, 4.24, 4.85, 5.22& \\
Turning angle(degree): 0.0, -3.08, -5.98, -4.66, -2.91, 1.97, 7.02, 14.46&\\
\VarSty{ {\bf Object boxes} } & \\
Frame 0: car:[0.298, 0.408, 0.572, 0.756], car:[0.924, 0.408, 1.0, 0.51], car:[0.005, 0.83, 0.995, 0.982], car:[0.737, 0.373, 0.933, 0.522], car:[0.737, 0.373, 0.933, 0.522]&\\
Frame 1: car:[0.327, 0.416, 0.623, 0.779], car:[0.004, 0.827, 0.99, 0.982], car:[0.961, 0.426, 1.0, 0.523], car:[0.76, 0.379, 0.966, 0.538]&\\
Frame 2: car:[0.393, 0.427, 0.709, 0.777], car:[0.79, 0.387, 0.945, 0.553], car:[0.003, 0.825, 0.99, 0.98], car:[0.926, 0.434, 1.0, 0.549]&\\
Frame 3: car:[0.518, 0.424, 0.849, 0.791], car:[0.834, 0.397, 0.994, 0.587], car:[0.007, 0.825, 0.983, 0.985]&\\
Frame 4: car:[0.695, 0.542, 0.924, 0.777]&\\
Frame 5: \\
Frame 6: \\ 
Frame 7: \\
    \hrulefill & \\
   \VarSty{ {\bf BDD-X QAs} } & \\
\textcolor{Q_rgb}{Question}: What is the  action of this vehicle?  &\\
\textcolor{A_rgb}{Answer}: The car turns right.& \\
\textcolor{Q_rgb}{Question}: Why does this vehicle behave in this way?  &\\
\textcolor{A_rgb}{Answer}: as the road is clear to turn.   &\\
\textcolor{Q_rgb}{Question}: Predict the speed of the ego vehicle. & \\
\textcolor{A_rgb}{Answer}: 5.5.& \\
\textcolor{Q_rgb}{Question}: Predict the turning angle of the ego vehicle. & \\
\textcolor{A_rgb}{Answer}:7.44.& \\
 \hrulefill & \\
   \VarSty{ {\bf QAs generated by ChatGPT} }& \\
\textcolor{Q_rgb}{Question}: What objects are present in the video, and how do they change throughout the frames?&\\

\textcolor{A_rgb}{Answer}: The video features various objects. As the video progresses, the positions and visibility of these objects change. In the initial frames, there are multiple cars. As the ego vehicle turns right, the surrounding cars gradually disappear from view. &\\



\textcolor{Q_rgb}{Question}: What can we learn from the ego vehicle's interactions with the traffic and surrounding environment?&\\

\textcolor{A_rgb}{Answer}: The ego vehicle's interactions with the traffic and surrounding environment demonstrate the importance of safe driving practices. The driver maintains a proper speed while navigating through traffic and pays attention to their surroundings, such as the positions of other vehicles. As they make the right turn, they ensure the road is clear before accelerating. These actions highlight the need for drivers to be aware of their environment, follow traffic rules, and prioritize safety when sharing the road with other users.&\\
    \end{tabular}
\end{mybox}
\caption{Example of the instruction-tuning data sample. The upper part of this figure demonstrates input information to ChatGPT, including video captions, control signals and object detection results obtained by YOLOv8. The lower part shows BDD-X QAs and conversations generated by ChatGPT. Refer to the appendix for detailed prompts.
}
    \label{tab:example_visual_instruction}
\end{minipage}

\end{table*}
\mypara{Additional QAs generated by ChatGPT.}\label{conv_qa_generate}
In previous works, ADAPT \cite{jin2023adapt} trains a caption network to predict descriptions and justifications. However, the provided description and justification labels are fixed and rigid. If human users wish to learn more about the vehicle and ask everyday questions, past works may fall short. Thus, BDD-X alone is insufficient for meeting the requirements of interpretable autonomous driving. Instruction tuning data generated by ChatGPT/GPT4 has been proven effective for performance enhancement in natural language processing \cite{peng2023instruction}, image understanding \cite{liu2023visual}, and video understanding \cite{li2023videochat,zhang2023video}. ChatGPT/GPT4 can access privileged information (e.g., image-labeled captions, ground truth object bounding boxes) and is prompted to generate conversations, descriptions, and reasoning. Currently, there is no visual instruction-following dataset tailored for autonomous driving purposes. Therefore, we create our own dataset based on BDD-X assisted by ChatGPT.

To address the aforementioned issue, ChatGPT is leveraged as a teacher to generate more conversations about the ego vehicle. The prompt generally follows the prompt design used in LLaVA. To enable ChatGPT to "see" the video, YOLOv8 \cite{reis2023real} is implemented to detect commonly seen objects in each frame of the video (e.g., vehicles, pedestrians). Obtained bounding box coordinates are normalized following LLaVA and sent to ChatGPT as privileged information. In addition to object detection results, the video clip's ground truth control signal sequences and captions are also accessible to ChatGPT. Based on this privileged information, ChatGPT is prompted to generate multiple rounds and types of conversations about the ego vehicle, traffic lights, turning directions, lane changes, surrounding objects, spatial relations between objects, etc. Detailed prompt is provided in the appendix.

Finally, we collect 56K video-text instruction-following samples, including 16K BDD-X QAs and 40K QAs generated by ChatGPT. An example is shown in Tab. \ref{tab:example_visual_instruction}.

\begin{table*}[t!]\centering
\begin{minipage}{2.0\columnwidth}\vspace{0mm}    \centering
\begin{mybox}{purple}
    \centering
   
      \footnotesize
    \begin{tabular}{p{0.97\columnwidth} c}

    \includegraphics[height=2cm]{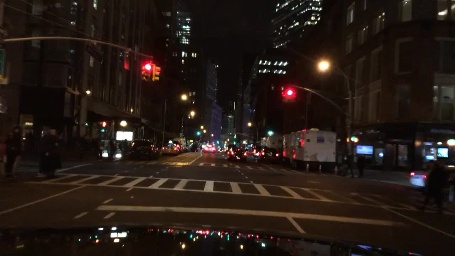}\hspace{3pt}\includegraphics[height=2cm]{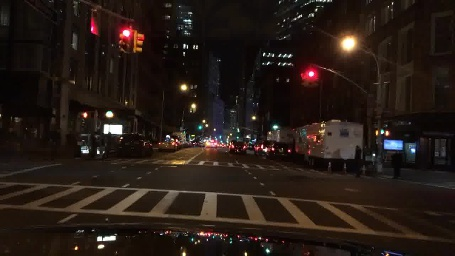}\hspace{3pt}\includegraphics[height=2cm]{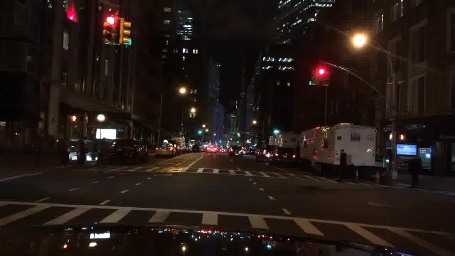}\hspace{3pt}\includegraphics[height=2cm]{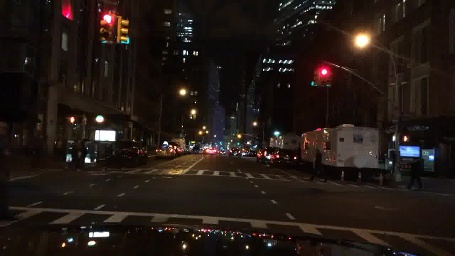}\\

\textcolor{Q_rgb}{Human}: Current speed of the vehicle: 2.15m/s. The length of this video is 2.10 seconds. What is the current action of this vehicle?
\textcolor{A_rgb}{Assistant}: The car slows down to a stop.\\

\textcolor{Q_rgb}{Human}: Why does the vehicle behave in this way?
\textcolor{A_rgb}{Assistant}: since the light ahead became red.\\
\textcolor{Q_rgb}{Human}: Forecast the speed of the vehicle.
\textcolor{A_rgb}{Assistant}: 2.09.\\
\textcolor{Q_rgb}{Human}: Forecast the turning angle of the vehicle.
\textcolor{A_rgb}{Assistant}: 0.00.\\
    \end{tabular}
\end{mybox}
\caption{Example of DriveGPT4 predictions. In this example, 4 out of 8 frames are shown for concise visualization.
}
    \label{tab:exmaple_input_output}
    \vspace{-15pt}
\end{minipage}

\end{table*}
\section{DriveGPT4}

\subsection{Model Architecture} 
DriveGPT4 is a versatile multimodal LLM capable of handling various input types, including videos, and texts. Videos are uniformly sampled into a fixed number of images, and a video tokenizer based on Valley \cite{luo2023valley} is employed to convert video frames into text domain tokens. All generated tokens are concatenated and input into the LLM. In this paper, LLaMA2 \cite{touvron2023llama2} is adopted as the LLM. Upon producing predicted tokens, a de-tokenizer decodes them to restore human languages. Drawing inspiration from RT-2 \cite{brohan2023rt}, texts and control signals utilize the same text de-tokenizer, signifying that control signals can be interpreted as a language and effectively processed by LLMs. Decoded texts contain predicted signals in a fixed format. The overview architecture of DriveGPT4 is visualized in Fig. \ref{diagram}. 

\mypara{Video tokenizer.} Let the input video frames be denoted as $V=[I_1,I_2,...,I_N]$. For each video frame $I_i$, the pretrained CLIP visual encoder \cite{radford2021learning} is used to extract its feature $F_i\in \mathbb{R}^{257\times d}$. The first channel of $F_i$ represents the global feature of $I_i$, while the other 256 channels correspond to patch features of $I_i$. For succinct representation, the global feature of $I_i$ is denoted as $F_i^G$, while the local patch features of $I_i$ are represented as $F_i^P$. The temporal visual feature of the entire video can then be expressed as:
\begin{equation}
    T = F_0^G \oplus F_1^G \oplus ... \oplus F_N^G 
\end{equation}
where $\oplus$ denotes concatenation. The spatial visual feature of the whole video is given by:
\begin{equation}
    S = \text{Pooling}(F_0^P, F_1^P,...,F_N^P)
\end{equation}
where Pooling$(\cdot)$ represents a pooling layer that convert $N$ features into a single feature tensor for memory efficiency. Ultimately, both the temporal feature $T$ and spatial feature $S$ are projected into the text domain using a projector.

\mypara{Text and control signals.}
Inspired by RT-2 \cite{brohan2023rt}, control signals are processed similarly to texts, as they belong to the same domain space.  Control signals are directly embedded within texts during the process. The default LLaMA tokenizer is employed. DriveGPT4 should predict control signals in the next step (i.e., $(v_{N+1},\Delta_{N+1})$) based on the multimodal input data. The speed of ego vehicle and time length of the input video clip are included in the text input. The turning angle represents the relative angle between the current frame and the previous frame. After obtaining predicted tokens, the LLaMA tokenizer is used to decode tokens back into texts.  Predicted control signals are embedded in the output texts using a fixed format, allowing for easy extraction.   An example illustrating the input and output of DriveGPT4 is presented in Tab. \ref{tab:exmaple_input_output}.

\subsection{Training}
Consistent with previous LLM-related studies, DriveGPT4's training consists of two stages: (1) the pretraining stage, focusing on video-text alignment; and (2) the mix-finetuning stage, aimed at training the LLM to answer questions related to interpretable end-to-end autonomous driving.

\mypara{Pretraining.}
In line with LLaVA \cite{liu2023visual} and Valley \cite{luo2023valley}, the model undergoes pretraining on 593K image-text pairs from the CC3M dataset and 703K video-text pairs from the WebVid-2M dataset \cite{Bain21}. The pretraining images and videos encompass various topics and are not specifically designed for autonomous driving applications. During this phase, the CLIP encoder and LLM weights remain fixed. Only the projector is trained.

\mypara{Mix-finetune.}
In this stage, the LLM in DriveGPT4 is trained alongside the projector. To enable DriveGPT4 to understand and process domain knowledge, it is trained with the 56K video-text instruction-following data generated in Section \ref{data_generation}. However the 56K autonomous driving domain data is not sufficient for LLM fine-tuning, and DriveGPT4 might have serious hallucination issues (e.g., detecting non-existent vehicles or traffic lights). To enhance DriveGPT4's ability for visual understanding and question answering, we scale up the fine-tuning dataset by utilizing 223K general instruction-following data generated by LLaVA and Valley for mix-finetuning. ``Mix'' represents that general visual understanding data is utilized for training together with task-specific instruction tuning data for our task. Consequently, DriveGPT4 is finetuned with 56K video-text instruction-following data for autonomous driving together with 223K general instruction-following data. The former ensures that DriveGPT4 can be applied for interpretable end-to-end autonomous driving, while the latter enhances the data diversity and visual understanding ability of DriveGPT4. For training efficiency, DriveGPT4 is first finetuned with 223K general data and then further finetuned by 56K domain specific data.  To further improve the reasoning ability of DriveGPT4 and handle the hallucination issue, in the future, we plan to create more instruction-tuning data based on the CARLA simulator.

\begin{table}[t]
  \caption{Testing set split.}
   
  \renewcommand\tabcolsep{4.3pt} 
  \small
  \label{testing_split}
  \centering
\begin{tabular}{lccccc}
\toprule
    Split
 & Scenes & Amount \\ 
    \midrule
    Easy & Stopped; Driving forward; Parked; etc.& 1202\\
    Medium & Lane switch; Acceleration; Intersection; etc. & 295 \\
    Hard & Vehicle turning; Traffic light change; etc. & 312\\
    \bottomrule
  \end{tabular}

\end{table}

\begin{table*}[t]
  \caption{Quantitative results of comparison experiments on different splits of the BDD-X testing dataset. We provide evaluation results on comprehensive text answering (i.e., combining description and justification). ``B4" represents the BLEU4 metric score.  }
  \renewcommand\tabcolsep{4.85pt} 
  \small
  \label{qa_compare_split}
  \centering
\begin{tabular}{lcccccccccccc}
\toprule
    \multirow{2}{*}{Method} & \multicolumn{3}{c}{Easy}            & \multicolumn{3}{c}{Medium}       & \multicolumn{3}{c}{Hard}  & \multicolumn{3}{c}{All}  \\ \cmidrule(l){2-4} \cmidrule(l){5-7} \cmidrule(l){8-10} \cmidrule(l){11-13} 
 & CIDEr$\uparrow$ &  B4$\uparrow$ & ROUGE$\uparrow$  & CIDEr$\uparrow$ &  B4$\uparrow$ & ROUGE$\uparrow$& CIDEr$\uparrow$ &  B4$\uparrow$ & ROUGE$\uparrow$ &CIDEr$\uparrow$ &  B4$\uparrow$ & ROUGE$\uparrow$\\ 
    \midrule
    ADAPT & 100.93& \textbf{20.90}  &46.17 & 62.66 &16.44 & \textbf{40.80}&  52.71 &\textbf{13.56} & 40.49&  85.38 &17.40 & 43.04 \\
    \midrule
    Video-LLaMA & 10.31 & 2.59 & 11.47 & 9.10 & 1.34 & 9.08 & 2.99 & 1.12 & 9.09 & 8.90 & 1.52 & 10.86   \\
    Valley & 31.27 & 5.31 & 44.29 & 17.35 & 4.10 & 31.55 & 9.76 & 2.34 & 20.46& 20.91 & 4.75 & 14.54\\
    \midrule
    DriveGPT4 & \textbf{113.20} & 20.38 & \textbf{46.46} & \textbf{65.01} & \textbf{16.94} & 40.51 & \textbf{57.29} & 12.28 & \textbf{42.07} & \textbf{99.10} & \textbf{18.32} & \textbf{44.73} \\
    
    \bottomrule
  \end{tabular}

\end{table*}

\begin{table*}[t]
  \caption{Quantitative results of comparison experiments on the whole BDD-X testing dataset. We provide evaluation results on action description, action justification, and full-text generation (i.e., i.e., combining description and justification). ``B4" stands for BLEU4.  }
  \renewcommand\tabcolsep{11.1pt} 
  \small
  \label{qa_compare}
  \centering
\begin{tabular}{lccccccccc}
\toprule
    \multirow{2}{*}{Method} & \multicolumn{3}{c}{Description}            & \multicolumn{3}{c}{Justification}       & \multicolumn{3}{c}{Full}  \\ \cmidrule(l){2-4} \cmidrule(l){5-7} \cmidrule(l){8-10} 
 & CIDEr$\uparrow$ &  B4$\uparrow$ & ROUGE$\uparrow$  & CIDEr$\uparrow$ &  B4$\uparrow$ & ROUGE$\uparrow$& CIDEr$\uparrow$ &  B4$\uparrow$ & ROUGE$\uparrow$ \\ 
    \midrule
    ADAPT & 227.93& 32.99  & 61.82 &  80.00 & 9.25 & 30.79 & 85.38 &17.40 &43.04\\
    \midrule
    DriveGPT4 &  \textbf{256.03}& \textbf{35.41} & \textbf{63.77} &  \textbf{98.71} &\textbf{10.02} & \textbf{31.52}&  \textbf{99.10} & \textbf{18.32} & \textbf{44.73}\\
    \bottomrule
  \end{tabular}

\end{table*}

\begin{table*}[!t]
  \caption{Quantitative results of 
control signals prediction on the whole BDD-X testing dataset.}
  \renewcommand\tabcolsep{10.1pt} 
  \small
  \label{control_compare}
  \centering
\begin{tabular}{lcccccccccc}
\toprule
    \multirow{2}{*}{Method} & \multicolumn{5}{c}{Speed (m/s)}            & \multicolumn{5}{c}{Turning angle (degree)}      \\ \cmidrule(l){2-6} \cmidrule(l){7-11}  
 & RMSE$\downarrow$  & $A_{0.1}\uparrow$  & $A_{0.5}\uparrow$ & $A_{1.0}\uparrow$ & $A_{5.0}\uparrow$  & RMSE$\downarrow$  & $A_{0.1}\uparrow$  & $A_{0.5}\uparrow$ & $A_{1.0}\uparrow$ & $A_{5.0}\uparrow$\\ 
    \midrule
    ADAPT & 3.02 & 9.56 & 24.77 & 37.07 & 90.39 & 11.98 & 27.93& 66.83 & 75.13 & 89.45\\
    \midrule
    DriveGPT4 & \textbf{1.30}& \textbf{30.09} &\textbf{60.88} &\textbf{79.92} &\textbf{98.44}& \textbf{8.98} &\textbf{59.23} &\textbf{72.89} &\textbf{79.59} &\textbf{95.32} \\
    \bottomrule
  \end{tabular}

\end{table*}

\begin{figure*}[!t]
  \centering
    \includegraphics[width=0.49\linewidth]{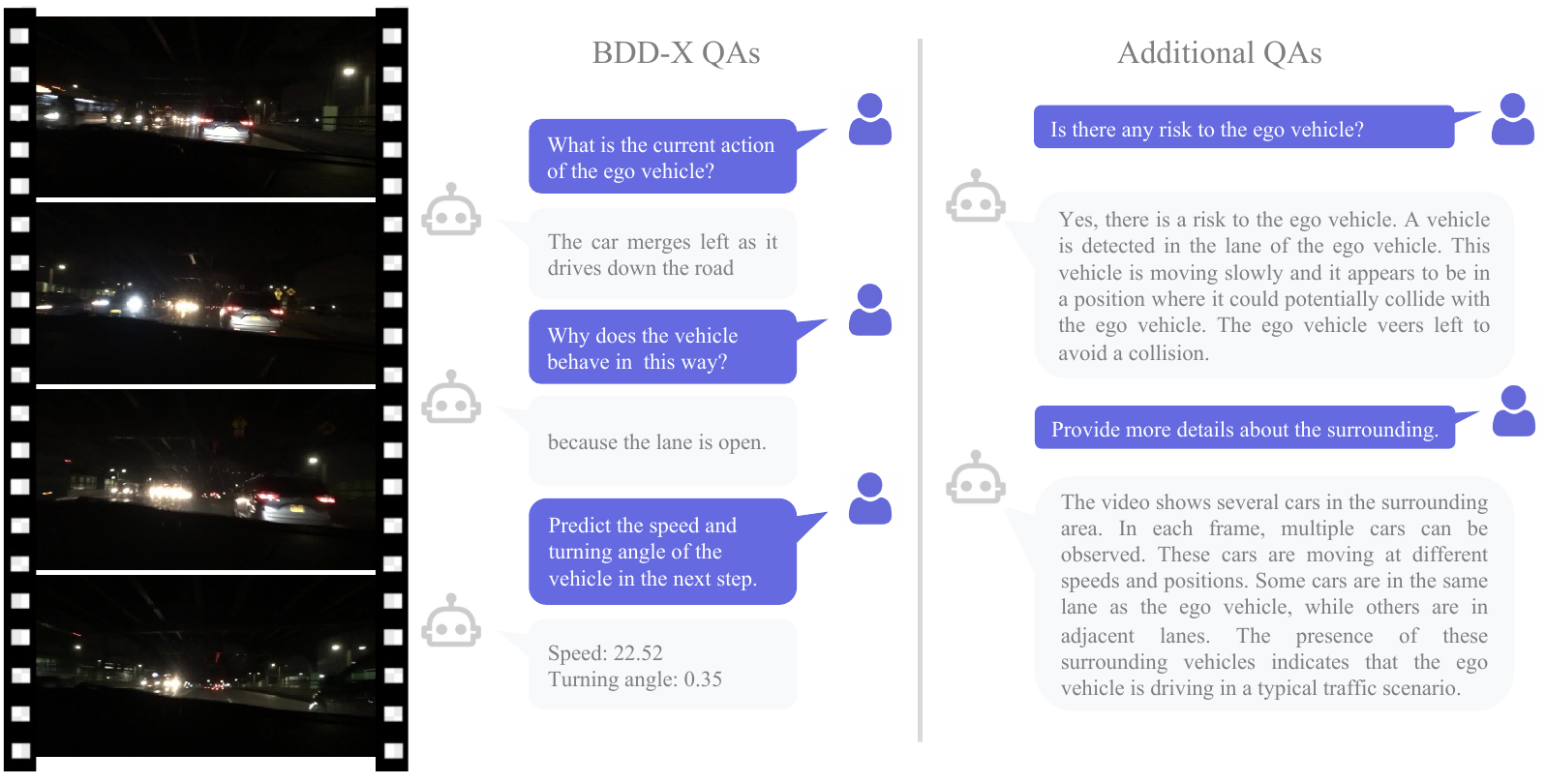}
     \includegraphics[width=0.49\linewidth]{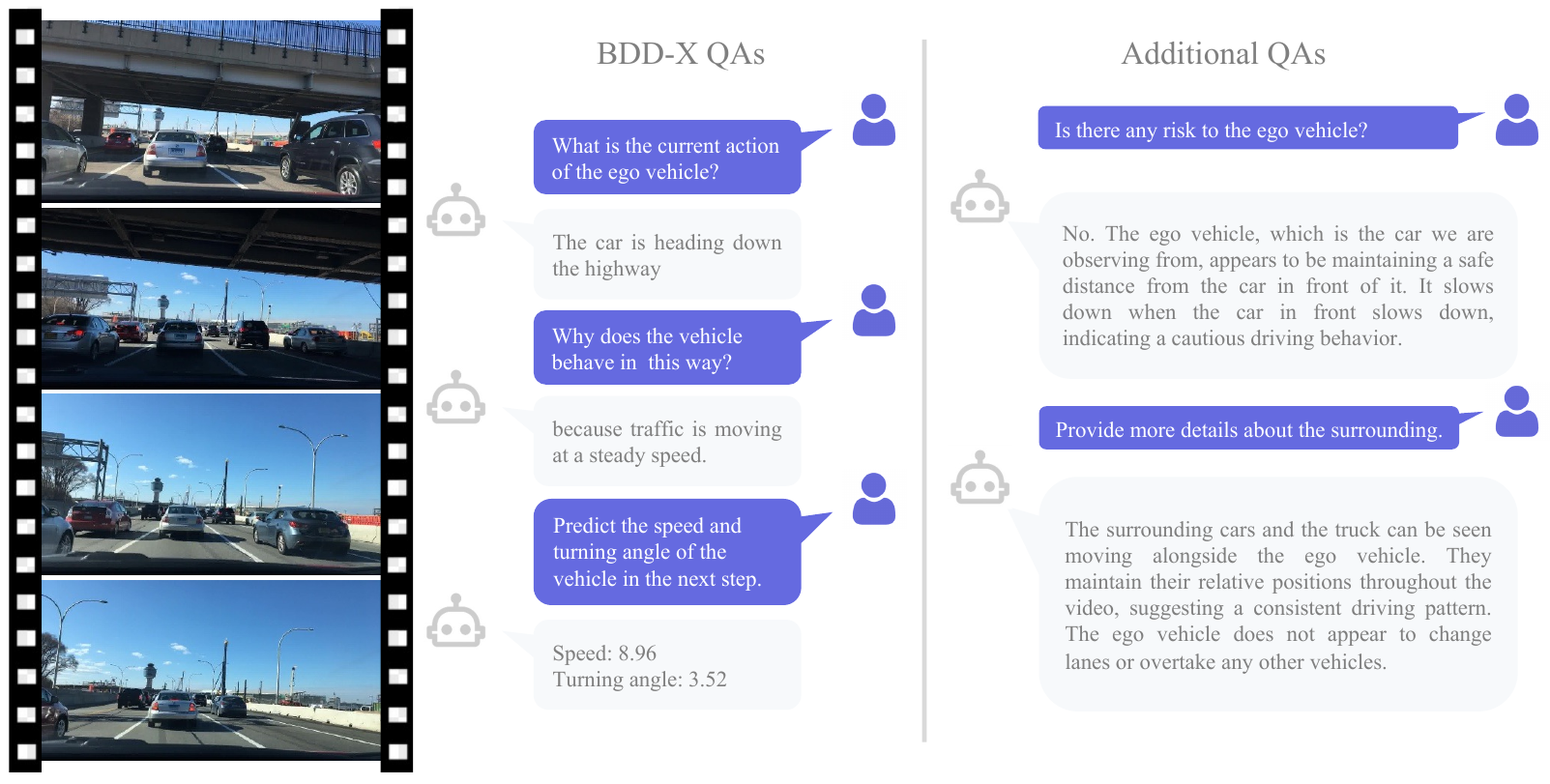}
  \caption{QAs of DriveGPT4 on the BDD-X testing set.}
  \label{example_1}
\vspace{-10pt}
\end{figure*}

\begin{table}[t]
 \vspace{-5pt}
  \caption{Quantitative results of comparison experiments on additional question answering. The model is required to answer questions generated by ChatGPT. ``B4" stands for BLEU4. ``-'' indicates the value is not available.}
   
  \renewcommand\tabcolsep{7.5pt} 
  \small
  \label{general_qa}
  \centering
\begin{tabular}{lccccc}
\toprule
    Method 
 &  CIDEr$\uparrow$  & B4$\uparrow$ &ROUGE$\uparrow$ & ChatGPT$\uparrow$ \\ 
    \midrule
    ADAPT & - & -& -& -\\
    \midrule
    Video-LLaMA & 5.71& 2.94 &10.20&27.75\\
    Valley &  11.37& 5.01 &11.09&43.23\\
    \midrule
    DriveGPT4  & \textbf{56.34} &\textbf{22.94} &\textbf{31.70} &\textbf{81.62}\\
    \bottomrule
  \end{tabular}
  \vspace{-15pt}

\end{table}

\begin{table*}[!h]
  \caption{Quantitative results of ablation studies on the BDD-X dataset. ``BQ", ``CQ", ``MF" represent BDD-X QAs, ChatGPT QAs and Mix-finetune, respectively. ``-'' indicates the value is not available.}
  \renewcommand\tabcolsep{5.06pt} 
  \small
  \label{ablation}
  \centering
\begin{tabular}{lccccccccccccccc}
\toprule
    \multirow{2}{*}{BQ}           &\multirow{2}{*}{CQ}           &\multirow{2}{*}{MF}
    &\multicolumn{3}{c}{BDD-X Questions}       & \multicolumn{3}{c}{ChatGPT Questions}            & \multicolumn{3}{c}{Speed}
    &\multicolumn{3}{c}{Turning Angle} \\ \cmidrule(l){4-6} \cmidrule(l){7-9} \cmidrule(l){10-12} \cmidrule(l){13-15} 
 &&& CIDEr$\uparrow$ & B4$\uparrow$  & ROUGE$\uparrow$  & CIDEr$\uparrow$ & B4$\uparrow$  & ChatGPT$\uparrow$ & RMSE$\downarrow$ & $A_{0.1}\uparrow$&$A_{1.0}\uparrow$&RMSE$\downarrow$ & $A_{0.1}\uparrow$&$A_{1.0}\uparrow$ \\ 
    \midrule
    &&& 20.91 & 4.75 & 14.54 & 11.37 &5.01 &  43.23 & - & - & - & - & -& -\\
    \checkmark&&\checkmark &95.75 &18.25 &44.03&   9.96&2.54 &31.03 & 1.69 & 22.82 & 77.91 & 9.97 & 55.97 & 76.11 \\
    &\checkmark&\checkmark  & 10.40& 2.31 &5.35 & 50.09 &21.53 &79.37 & - & - & - & -& -&- \\
    \checkmark&\checkmark& &76.51 & 11.09 & 40.93 & 17.24 & 10.50 & 77.37 & 4.67 & 17.15 & 44.82 & 14.80 & 21.19 & 70.02 \\
    \checkmark&\checkmark& \checkmark& \textbf{99.10}& \textbf{18.32}& \textbf{44.73}& \textbf{56.34} & \textbf{22.94} & \textbf{81.62} & \textbf{1.30} & \textbf{30.09}& \textbf{79.92} & \textbf{8.98} & \textbf{59.23}&\textbf{79.59} \\
    \bottomrule
  \end{tabular}
\end{table*}

\section{Experiment}
In this paper, DriveGPT4 focuses on interpretable end-to-end autonomous driving. With video frames and human questions as input, the method is required to predict interpretations in human language and control signals in the next step. Currently, except the BDD-X dataset, there are very few existing datasets that provide video clips captured by vehicle-mounted cameras with text interpretation and control signal annotations. Therefore, we mainly conduct evaluation experiments on the BDD-X dataset. The BDD-X dataset is filtered to remove samples that have inconsistent control signals and text reasoning. 

\subsection{Interpretable Autonomous Driving}
In this section, we evaluate DriveGPT4 and its baselines on interpretation generation, covering vehicle action description, action justification, and additional questions about vehicle status. ADAPT \cite{jin2023adapt} serves as the state-of-the-art baseline work. Recent multimodal video understanding LLMs \cite{zhang2023video,luo2023valley} are also considered for comparison. All methods use 8-frame videos as input. Currently, DriveGPT4 does not take 32-frame videos as input like ADAPT considering the heavy memory consumption and inference speed, which could be treated as a limitation of this work.

\mypara{Testing Set Split.} 
During vehicle driving, the distribution of scenes is usually not balanced. For example, some simple scenes like driving straight-forward are more commonly seen than more challenging vehicle turning or lane changes. For a comprehensive evaluation comparison, the testing set is split into ``Easy", ``Medium" and ``Hard" sets based on the driving scene and vehicle status. Detailed split information is shown in Tab. \ref{testing_split}.

\mypara{Evaluation Metrics.} To thoroughly assess the methods, we report multiple metric scores widely used in the NLP community, including CIDEr \cite{vedantam2015cider}, BLEU4 \cite{papineni2002bleu}, and ROUGE-L \cite{lin2004rouge}. The BDD-X QA task tends to have a fixed format, so the aforementioned NLP metrics are already sufficient for evaluation. ChatGPT-generated QAs possess flexible formats and more complicated semantic meanings. Following past MLLM works \cite{liu2023visual,li2023videochat,luo2023valley}, we also report the score generated by ChatGPT. ChatGPT is prompted to assign a numerical score between 0 and 1, with a higher score indicating better prediction accuracy. The detailed prompt for ChatGPT-based evaluation is available in the appendix. However, it should be noted that the ChatGPT score is not stable, thus we report the mean of three times of evaluations for reference.

\mypara{Action Description and Justification.} 
The goal is to predict vehicle action descriptions and justifications as closely as possible to the given labels. Evaluation results of all testing splits are displayed in Tab. \ref{qa_compare_split}. More detailed results are shown in Tab. \ref{qa_compare}. From the results, it is observed that DriveGPT4 outperforms the previous SOTA baseline ADAPT on all testing data, especially for the ``Hard" splits with more challenging driving scenes and vehicle dynamics. The effectiveness and superiority of the proposed DriveGPT4 are well demonstrated.

\mypara{Additional Question Answering.}
The above vehicle action description and justification have relatively fixed formats. To further evaluate the interpretable ability and flexibility of DriveGPT, additional questions are generated following section \ref{conv_qa_generate}. 
A hundred randomly sampled video clips in the BDD-X testing set are used for question generation. Compared with action descriptions and justifications, these questions are more diverse and flexible. The evaluation results are shown in Tab. \ref{general_qa}. ADAPT cannot answer additional questions except for the vehicle action description and justification. Previous video understanding LLMs can answer these questions but they do not learn autonomous driving domain knowledge. Compared with all baselines, DriveGPT4 presents superior results, demonstrating its flexibility.


\subsection{End-to-end Control}
In this section, we evaluate DriveGPT4 and its baselines for open-loop control signal prediction, specifically focusing on speed and turning angle. All methods are required to predict control signals for the next time step. Following previous works on control signal prediction, we use root mean squared error (RMSE) and threshold accuracies ($A_\tau$) for evaluation. $A_\tau$ measures the proportion of test samples with prediction errors lower than $\tau$. For a comprehensive comparison, we set $\tau$ with multiple values: $\{0.1,0.5,1.0,5.0\}$. The quantitative results for the previous state-of-the-art (SOTA) method ADAPT and DriveGPT4 are shown in Tab. \ref{control_compare}. DriveGPT4 achieves superior results for both speed and turning angle predictions.

\subsection{Qualitative Results.}

\begin{figure}[!t]
  \centering
    \includegraphics[width=\linewidth]{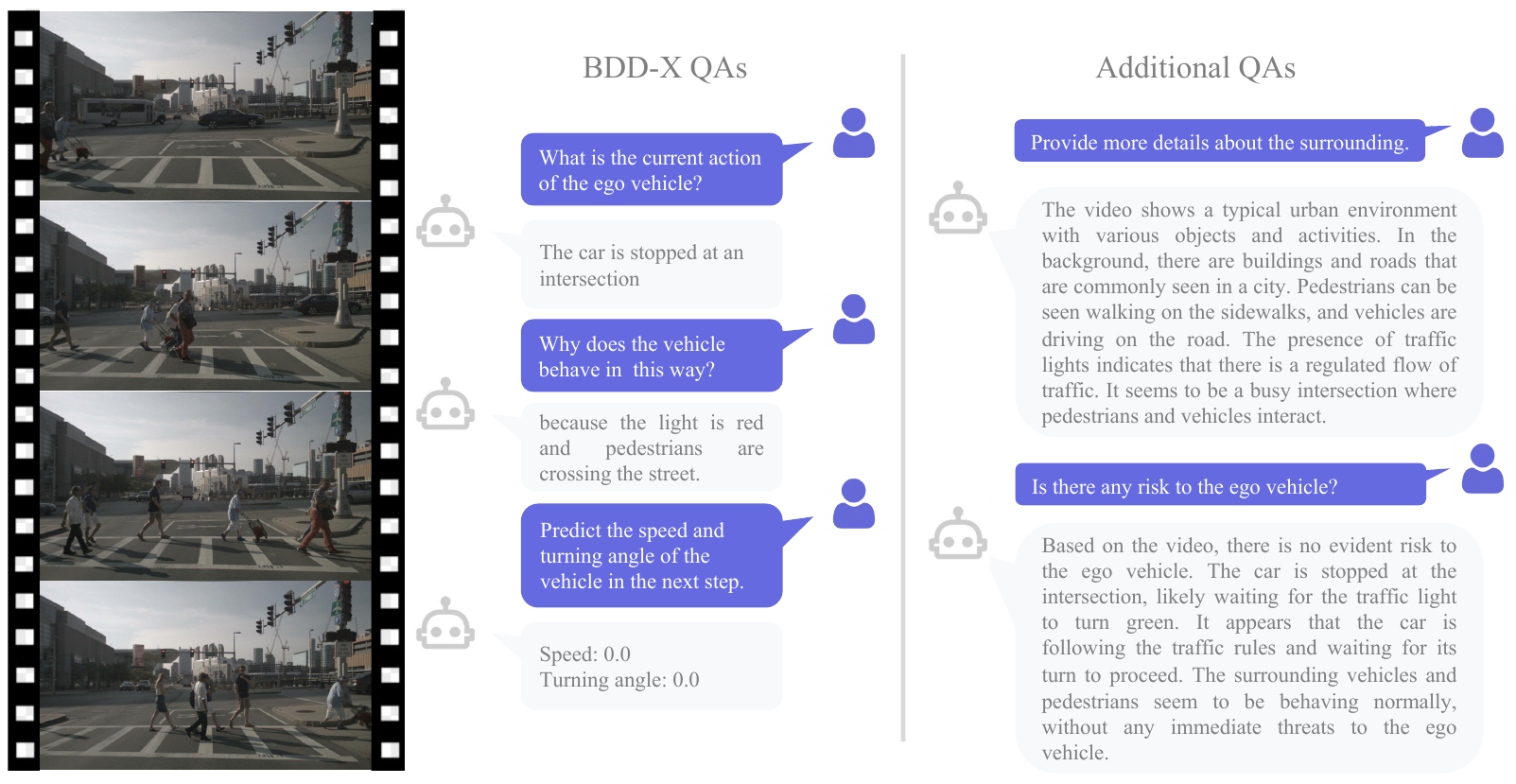}
  \caption{Zeroshot generalization of DriveGPT4 on NuScenes \cite{caesar2020nuscenes}.}
  \label{example_nusc}
\end{figure}

\begin{figure}[t]
  \centering
    \includegraphics[width=\linewidth]{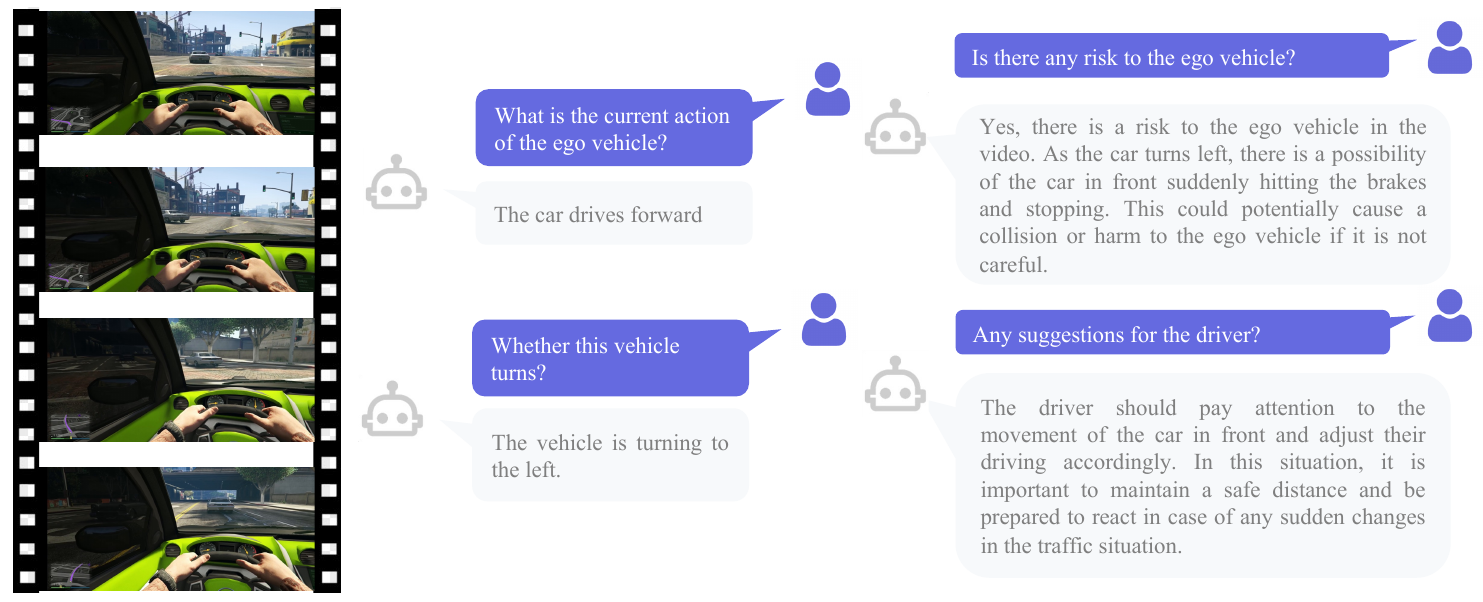}
    \vspace{-15pt}
  \caption{Zero-shot generalization of DriveGPT4 on video games.}
  \label{example_gta}
\end{figure}

Multiple qualitative results are provided for intuitive comparison. For concise visualization, we only show four frames of the input video clip. First, an example from the BDD-X testing set is visualized in Fig. \ref{example_1}. DriveGPT4 can generate high-quality texts and control predictions based on the prompt. Then, to verify the generalization ability of DriveGPT4, we apply DriveGPT4 to the NuScenes dataset \cite{caesar2020nuscenes} for zero-shot QA in Fig. \ref{example_nusc}. We also try DriveGPT4 on video games to further test its generalization ability. An example is shown in Fig. \ref{example_gta}.

\begin{figure}[t]
  \centering
    \includegraphics[width=\linewidth]{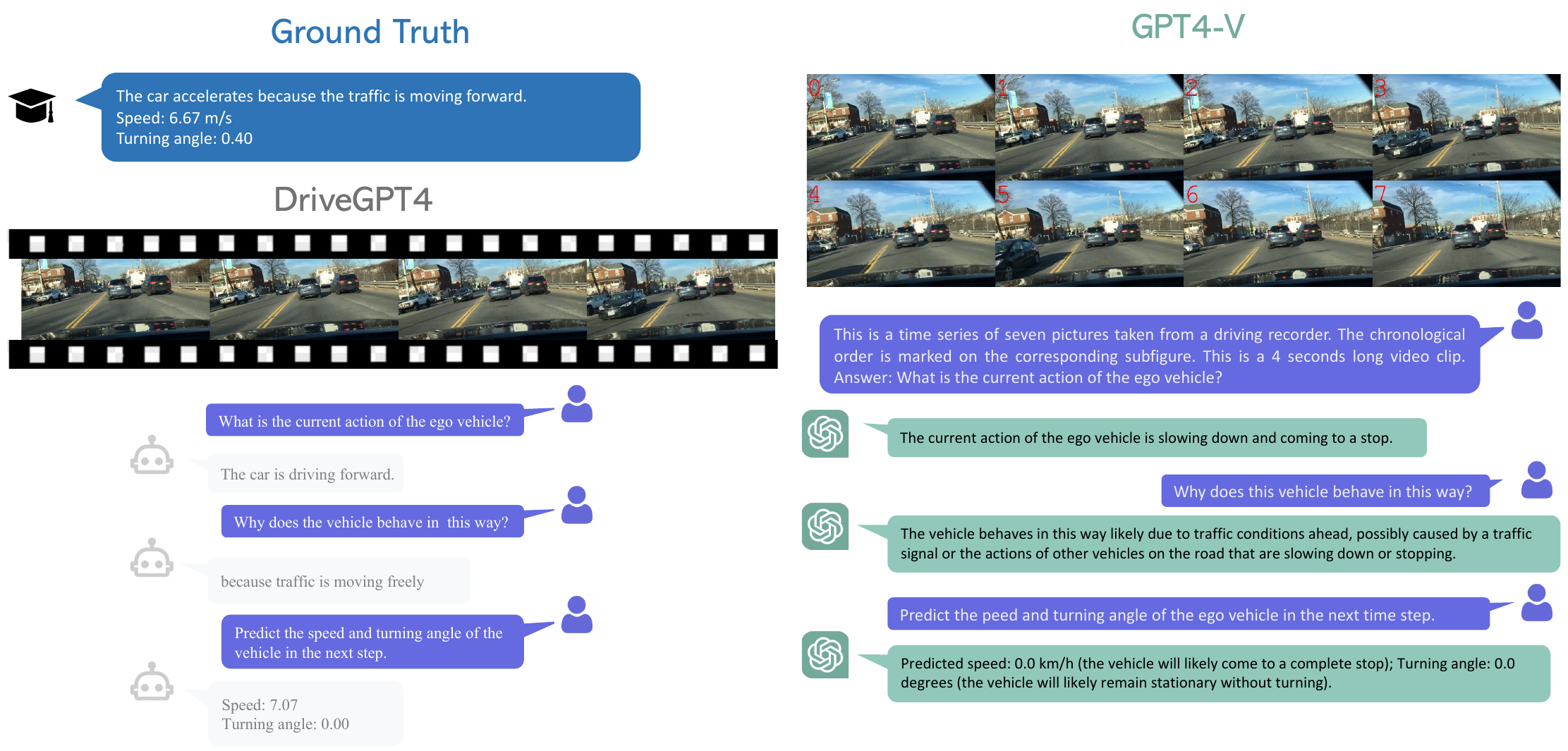}
  \caption{Comparison of DriveGPT4 and GPT4-V. GPT4-V is prompted with BDD-X QA pairs before the comparison.}
  \label{example_gpt4-v}
\end{figure}

\mypara{GPT4-V.}
As the multimodal version of GPT4, GPT4-V can understand, and reason single-frame images, illustrating excellent generalization ability for various daily tasks. However, GPT4-V is still a general model for images, and not specially finetuned for grounding autonomous driving applications. Before the comparison, GPT4-V is prompted with several BDD-X QA pairs in advance. During the qualitative evaluation, even though GPT4-V illustrates powerful recognition and reasoning ability, it is observed that it (1) cannot predict numerical control signals; (2) fails to correctly understand some vehicle actions, especially dynamic actions (e.g., turning, accelerating, etc.). An example is shown in Fig. \ref{example_gpt4-v}. More examples can be found in the appendix.

\subsection{Ablation Studies}
In this paper, several ablation studies are conducted to validate proposed designs, and the results are provided in Tab. \ref{ablation}.  By removing either BDD-X QAs or ChatGPT QAs during finetuning, a decrease in corresponding performance is observed, highlighting the significance of including all task-specific multimodal data. QA pairs generated by ChatGPT enable DriveGPT4 to answer human questions in more flexible patterns, and enhance the QA ability of BDD-X questions. Then, we test DriveGPT4 without the mix-finetune strategy by removing the general image and video instruction-following data. Severe performance deduction is observed, indicating the necessity of finetuning DriveGPT4 with diverse multimodal data. Thus, changes to DriveGPT4 would negatively impact its versatile QA capabilities for interpretable end-to-end autonomous driving.

\section{Conclusion}
This paper presents DriveGPT4, an interpretable end-to-end autonomous driving system using multimodal LLM. A new dataset for autonomous driving interpretation is developed with the assistance of ChatGPT and employed to mix-finetune DriveGPT4, enabling it to respond to human inquiries about the vehicle. DriveGPT4 utilizes input videos and texts to generate textual responses to questions and predict control signals for vehicle operation. It outperforms baseline models in various tasks such as vehicle action description, action justification, general question answering, and control signal prediction. Moreover, DriveGPT4 exhibits generalization ability through zero-shot adaptation. In the future, DriveGPT4 will be further enhanced for close-loop vehicle control tasks. To handle the drifting issue of imitation learning, an LLM expert will be developed for data collection without human effort. 


\bibliographystyle{IEEEtran}
\bibliography{main}

\end{document}